\documentclass{article}
\usepackage{graphicx} 
\usepackage{comment}
\usepackage{hyperref}

\title{
Fantastic Biases (What are They) and Where to Find Them}
\author{Valentin Barriere\\
Universidad de Chile -- DCC -- CENIA\\
Beauchef 851, Santiago, Chile\\
\texttt{vbarriere@dcc.uchile.cl}
}
\date{June 2024}

\usepackage{acl}

\begin{document}

\maketitle

\begin{abstract}
Deep Learning models tend to learn correlations of patterns on huge datasets. The bigger these systems are, the more complex are the phenomena they can detect, and the more data they need for this. 
The use of Artificial Intelligence (AI) is becoming increasingly ubiquitous in our society, and its impact is growing everyday. The promises it holds 
strongly depend on their fair and universal use, 
such as access to information or education for all. In a world of inequalities, they can help to reach the most disadvantaged areas. 
However, such a universal systems must be able to represent society, without benefiting some at the expense of others. We must not reproduce the inequalities observed throughout the world, but educate these IAs to go beyond them. 
We have seen cases where these systems use gender, race, or even class information in ways that are not appropriate for resolving their tasks. Instead of real causal reasoning, they rely on spurious correlations, which is what we usually call a bias. 
In this paper, we first attempt to define what is a bias in general terms. It helps us to demystify the concept of bias, to understand why we can find them everywhere and why they are sometimes useful. Second, we focus over the notion of what is generally seen as negative bias, the one we want to avoid in machine learning, before presenting a general zoology containing the most common of these biases. 
We finally conclude by looking at classical methods to detect them, by means of specially crafted datasets of templates and specific algorithms, and also classical methods to mitigate them. \footnote{This pre-print is an English version of a paper published in Spanish \cite{BarriereBits24}.}

\end{abstract}

\section{Are we Talking About Fairness?}

\begin{figure}
    \centering
    \includegraphics[width=.4\textwidth]{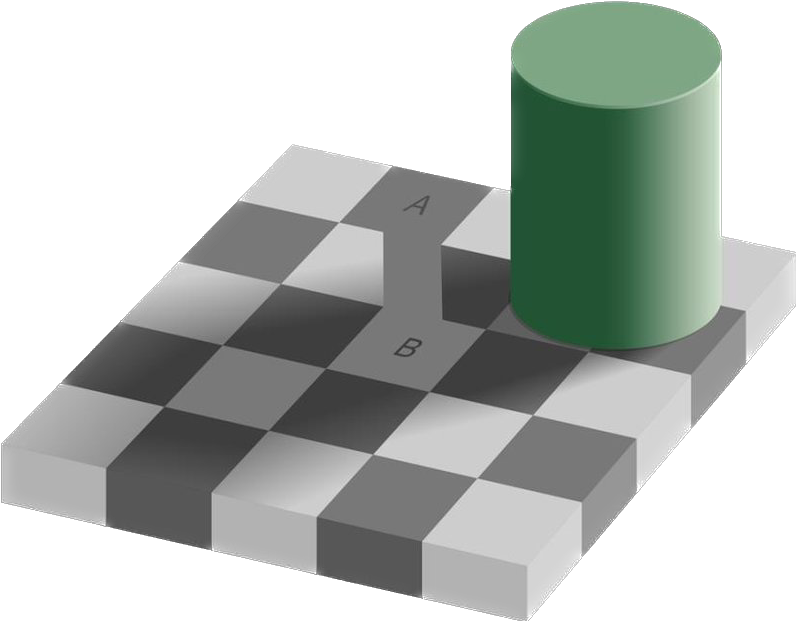}
    \caption{The chessboard shadow illusion}
    \label{fig:chess}
\end{figure}

\subsection{In the society...}

Fairness in Artificial Intelligence (AI) models is a huge concern for society today. 
Biased algorithms are shaking up society in ways we never imagined, often reinforcing existing inequalities and discrimination. Imagine an algorithm deciding who gets a loan, who gets hired, or even who gets bail. If these algorithms are biased, they can unfairly target certain groups, like minority communities or women, perpetuating injustice and discrimination.

In criminal justice, biased algorithms can lead to higher incarceration rates for minorities, while in hiring, they might favor men over equally qualified women. Healthcare is not immune either—biased algorithms can result in poorer treatment recommendations for underrepresented groups, with dire consequences. The ethical implications of these biases extend beyond direct discrimination to issues of accountability and transparency. These algorithms often operate as black boxes, making their decision-making processes opaque and unchallengeable. This lack of transparency erodes trust and raises serious accountability issues. Who’s responsible when an algorithm discriminates?

The answer isn’t simple, but it starts with diversifying the data, implementing bias detection techniques, and fostering collaboration among technologists, ethicists, and policymakers. Raising public awareness about these digital biases is crucial. While algorithms promise efficiency and innovation, their ethical deployment must prioritize fairness and justice to ensure they benefit everyone equally without deepening existing societal divides. 

As AI becomes increasingly pervasive, the quest for high performance drives models to become more complex, often relying on correlational associations. Yet, we also demand that these models exhibit unbiased behavior, respecting diversity and relying on causation rather than correlation. 
%
However, let's not forget that biases are everywhere, as almost nothing in the world is pure randomness, structures are everywhere. 

Mathematics can be defined as the study of structures, but similar structures are sharing a common structural bias, which can complicate the effort to use them without perpetuating unfair correlations. This creates a significant dilemma: how can we leverage the useful priors of the world to make probable decisions without falling into the trap of non-causal biased correlations that can be harmful in some sensitive cases? 
For example, the crime rate is higher is certain subgroups in the population, a probabilistic pick is not a causal reason to know that an individual will commit a crime even though it might be statistically probable regarding hypotheses. \citet{ObservatoiredesInegalites2021} reported that Black and Arabic young men, which are subgroups of the population with a higher crime rate, are more prone to be controlled by the police. It is a common practice called \textit{racial profiling}, which sanctions based on social and ethnics prejudices, and leads to an unfair treatment and systemic bias in law enforcement practices \citep{Jounin2015}. 
%
The challenge lies in removing such detrimental biases from AI models without sacrificing their ability to make informed, probabilistic decisions based on the world's inherent structures. In other words: \textbf{how to remove detrimental biases in AI models?}   


\subsection{...and in AI models}

Ensuring fairness in machine learning (ML) is a complex and challenging task. Firstly, ML models are trained on real-world data, which inherently contains biases related to race, gender, religion, social class, or whatever. As a result, these models can not only learn but also amplify these pre-existing biases \citep{Hall2022}, leading to problematic outcomes. 

Secondly, despite rigorous and comprehensive training and testing, creating systems that behave fairly across all situations and cultures remains a significant challenge \citep{Mehrabi2021,Santy2023}. For instance, consider a chatbot trained on Castilian Spanish. It might perform well when interacting with users from Spain, but might struggle to understand and respond appropriately to evolving slang used by teenagers. When used by speakers from Chile or Argentina, the chatbot could have difficulty adapting to the unique idioms and new polysemy of those regions, resulting in miscommunication and frustration (or laughter, but poor performances). Even within its initial context, such as among Castilian speakers, the chatbot might encounter problems with specific dialects or local events, revealing unexpected and unfair outcomes post-deployment.

Thirdly, defining fairness is inherently complex, as there is no universally accepted standard, whether for human or machine decisions. Determining appropriate fairness criteria for a system requires balancing user experience, cultural, social, historical, political, legal, and ethical considerations, each with potential trade-offs. Even in seemingly straightforward situations, there can be disagreement over what constitutes fairness, as it can rely on the values of the individuals \cite{Sorensen2023,Mirzakhmedova202}, complicating the establishment of AI policies, especially in a global context. 



Finally, generative AI is becoming an increasingly important part of our lives, and generates more and more content on the internet, which is the primary source of data for training new models. 
The inherent bias of contemporary generative models is a poison for future generative models, as it will lower the quality of the future training data \cite{Taori2023}.

Nonetheless, striving for continuous improvement towards "fairer" systems is essential, and even if absolute fairness remains elusive, there are ways to tend to it by removing the biases of existing models. 
But before talking about debiasing IA, let's start simple: what is a bias? We will show in the following that \textit{bias} have many senses, starting from its various definitions and uses in the language.

\section{What are biases?}

\subsection{General Definition of biases}

First, let’s think about the mathematical bias of a linear model. This is the value that a model will output when the inputs are zero. A model is then biased when it outputs a value different than zero when transforming the vector null. 
It is the deviation that the model introduces from the zero element. 
This bias helps the model to fit the data, because every data is biased. 

Second, let’s think about cognitive biases. These are biases that all humans are sharing. 
They can be seen as a difference between the reality and our perception of the world. They can be very low-level like a simple optical illusion,\footnote{which acts very much likes cognitive bias \cite{Baer2017}} or of higher-level of complexity, constructed on social concepts. At lower-level of complexity, you can see an example in Figure \ref{fig:chess} a chessboard  with a cylinder on it: the dark squares outside the shadow looks darker than the white squares in the shadow, even though in reality they have the same dark grey color. 
Another case where we can separate groups of people regarding their bias is the famous gold and white vs black and blue dress.\footnote{If you need a reminder, it's right \href{https://upload.wikimedia.org/wikipedia/en/2/21/The\_dress\_blueblackwhitegold.jpg}{here}}
 
\begin{figure}
    \centering
    \includegraphics[width=.5\textwidth]{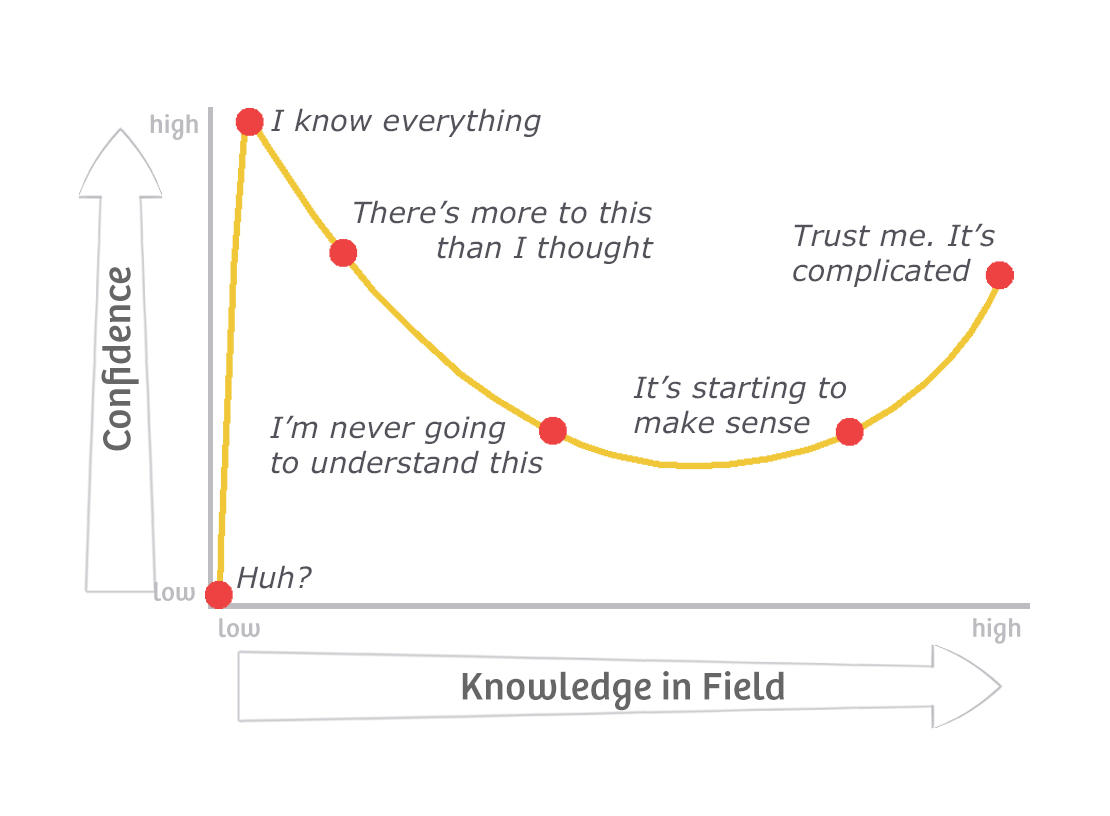}
    \caption{Duning-Krugger effect}
    \label{fig:dk_effect}
\end{figure}

At higher-level of complexity, we can mention among others the Dunning-Kruger effect, the availability bias, or the confirmation bias. The Dunning-Kruger effect, illustrated in Figure \ref{fig:dk_effect}, is a the tendency for an individual with limited knowledge or competence in a given field to overestimate their own skills in that field.\footnote{Confirmation bias is the tendency for the brain to value new information that supports existing ideas, and availability bias is the tendency for the brain to conclude that a known instance is more representative of the whole than is actually the case.} One of the greatest popularization book about human cognitive biases is "Thinking fast and slow" from \citet{kahneman2011thinking}, who won a Nobel Prize in Economy for his life work on psychology of judgment and decision-making. These biases can be seen as a deviation between our perception and the reality. 

Third, we can name the social biases, like social norms which are cultural biases. It can be as simple as the proper way to dress yourself, but also involves behavioral aspects. The expectations from people of different social groups will vary with respect to the group, depending on what it is expecting from one of its members. 
When two individuals from different cultures do not know the social norms of the other ones, they can experience what is called a culture clash. Culture clash is defined by Jonathan H. Turner\footnote{Professor of sociology at University of California, Riverside} as "differences in cultural values and beliefs that place people at odds with one another". In some culture, saying "no" can be seen as a sign of weakness, when in some others as a sign of strength. 
Note that social norms depend on social groups, so you can even find differences within one culture, norms can change regarding the social group: you cannot have the same kind of behavior as a man or as a woman in many countries.  

Biases are not fundamentally bad, they just are a deviation from a (subjective and defined) norm or value. For example, cognitive biases come from the structure of our brain, which is not random. We all share a significant portion of our DNA, which could be seen as unbiased if it was center of zero: a random sequence of nucleotides,\footnote{A, T, G and C} like a 
distribution centered over zero. 
Differences in hair, eye color, skin tone, nose or mouth shape can reflect phenotypic variations. Phenotypes include similar patterns in DNA: again, 
individuals from the same subgroup share the same biases when compared to outside subgroups. 

We rely on biases every time we take a decision, it is terribly useful in order to select the most probable choices. 
In their ACL paper, \citet{Meister2022} show that an intuitive bias for a language model\footnote{A model supposed to predict next word knowing the precedent ones, such as what you have when written a text on your smartphone} is the actual use frequency of the different words. It reflects the way of speaking, which is a bias: you know that an Argentinean and Chilean will not use the same words. For this, they simply initialized the bias term of the language model’s final linear layer with the log-unigram distribution of the words. 
This helps using \textit{prior knowledge} to minimize the loss in a very naive way, and links the two notions of bias we introduced before! 




\subsection{Zoology of common biases in science}

It exists a complete zoology of biases, they come from a gap between the perception and the truth. They can \textit{in fine} impact the validity and reliability of findings which lead to misinterpretation of data. 
More biases are described \href{https://www.scribbr.com/category/research-bias/}{in this article}.

\subsubsection*{Reporting bias} 
This bias occurs when the frequency of events, properties, and outcomes recorded in a dataset does not accurately reflect their real-world prevalence. It often arises because people tend to document unusual or memorable circumstances, assuming that ordinary events are not worth noting: the phenomenon exists but people do not report it. At another level, it can result from the tendency to publish only successful experiments or positive results, which creates a skewed perception of a model’s effectiveness and mislead about its true capabilities and limitations.

\subsubsection*{Automation bias} 
It is the tendency to prefer results produced by automated systems over those produced by non-automated systems, regardless of the respective error rates. 
That is how you arrive to a lousy chatbot not understanding a user's special request, which should be handled by a human.

\begin{figure}
    \centering
    \includegraphics[width=.4\textwidth]{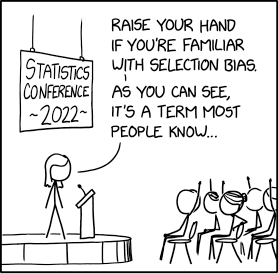}
    \caption{An example of coverage bias from the xkcd comics \href{https://xkcd.com/2618/}{\#2618}}
    \label{fig:selection_bias}
\end{figure}

\subsubsection*{Selection bias} 
It occurs if a dataset's examples are chosen in a way that is not reflective of their real-world distribution.\footnote{It is different than reporting bias} This bias can take various forms. The \textbf{coverage bias} arises when some groups are inadequately represented in the training data, like surveying in a Computer Science classroom in order to poll about how much people know about programming (see Figure \ref{fig:selection_bias}). The \textbf{participation bias} occurs when only the people interested will answer a study and the \textbf{sampling bias} is due to a non-randomization of the answers (or of the data during the training!).  

\subsubsection*{Representation Bias} 
This bias is a bit similar to the selection bias but with a subtlety. It happens when the data collected only represent a subgroup of the population, even though it represents the reality. The fact that mostly males are CEOs does not mean that gender is a feature to success as a CEO.

\subsubsection*{Group attribution bias} 
It involves overgeneralizing characteristics, based on limited observations of individuals, to the entire group to which they belong. 
It can be an \textbf{in-group bias}, which tend to favor the individuals from the same group as the experimenter, or \textbf{out-group homogeneity bias} which will tend perceiving members of an out-group as more similar to each other than they actually are, like a Castilian-speaking data scientist making one category for Castilian-Spanish and another for the rest of the variations. 

\subsubsection*{Implicit bias} 
This bias occurs when assumptions are made based on one's own mental models and personal experiences that do not necessarily apply more generally. It is called a \textbf{confirmation bias} when one processes data in ways that affirm preexisting beliefs and hypotheses, such as discarding out-of-distribution without cause, or \textbf{experimenter's bias} when one conditions the experiment to reach the expected conclusion.\footnote{An example of experimenter's bias is depicted \href{https://xkcd.com/882/}{\underline{here}}.} 

\subsection{Example of Biases in ML, NLP and LLMs}

Nowadays models need way more data to train, but available data represents what is on the internet, which does not necessarily represent the real world. 
And even if it accurately reflects real-life distribution, there's no guarantee it isn't biased, as the real world itself contains inherent biases. 
Indeed, with skewed distributions the model will tend to overfit on specific spurious characteristics: here come the biases. 

In natural language processing (NLP), biases are everywhere, starting by the data \cite{Wiegand2019}, the annotations \cite{Santy2023,Sap2022} and even the annotation campaign instructions \cite{Parmar2023}. Among other things, NLP models can drag moral \cite{Hammerl2022}, social \cite{Sap2020} or political biases \cite{Feng2023}. 
The quantification of social bias is a prominent theme in recent research. It can be in multimodal data like image captioning \cite{Hirota2022} or just in general text \cite{Czarnowska2021}. 

\begin{figure}
    \centering
    \includegraphics[width=.45\textwidth]{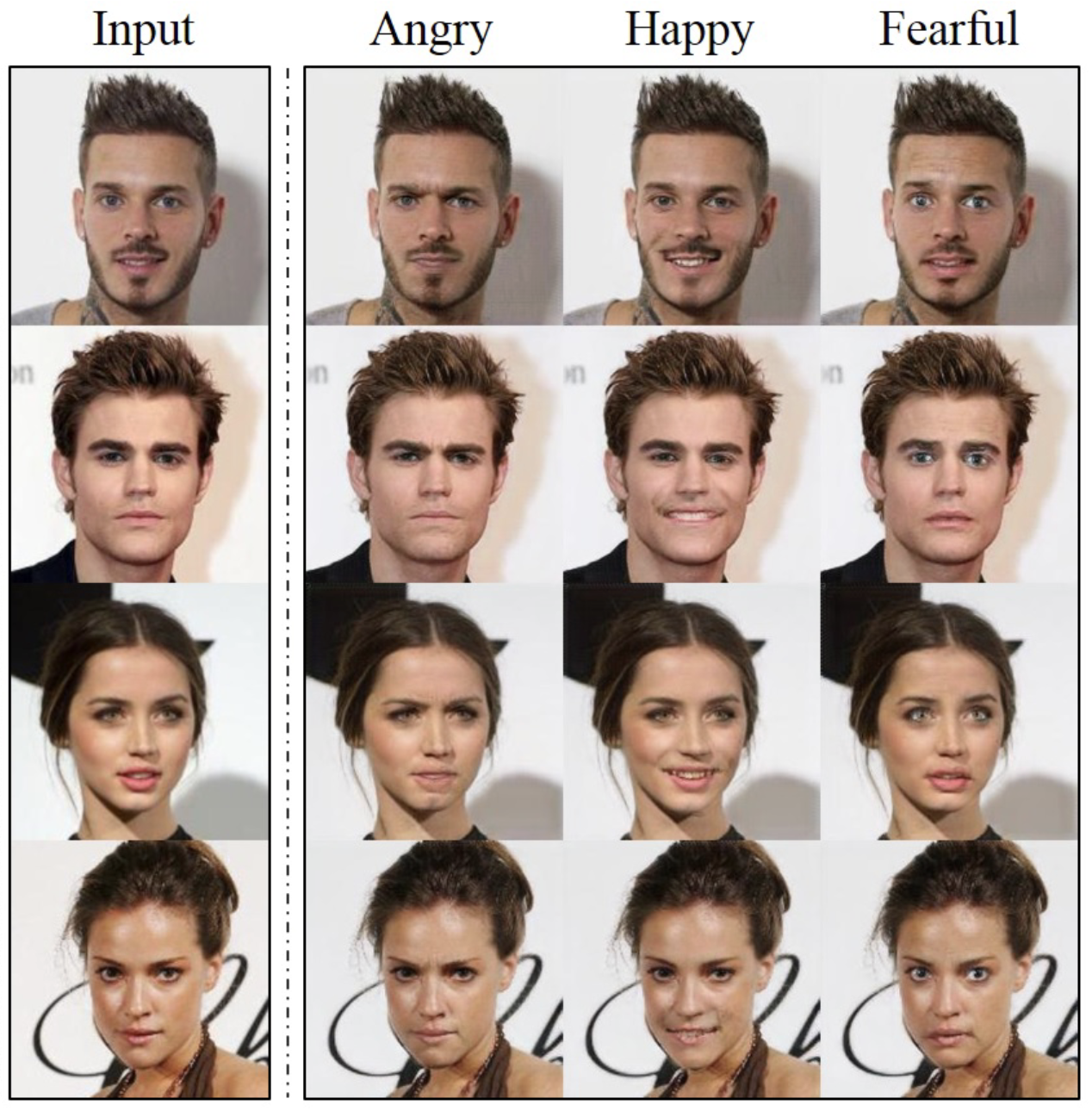}
    \caption{The Stargan model, trained over datasets containing mainly white young faces such as CelebA}
    \label{fig:stargan}
\end{figure}

In general, the main biases comes from a \textbf{Selection Biases}. The main part of the internet data is Western-centered. Generative models like StarGAN (\citet{Choi2018}; see Figure \ref{fig:stargan}) were mainly creating faces of white people because of its training set (CelebA from \citet{Liua2015}). This still stands with new models such as Stable Diffusion or Dall-E2. Even though now these models are more diverse, they create stereotypes: generating images of head-covered muslim men, female cleaning employee but productive persons as white men, and dark-skin colored persons for someone accessing social services \cite{Tiku2023}. 
This still remains the main source of bias for Large Language Models today.

\subsubsection*{Feature selection bias}
In old-school machine learning algorithms, selection bias at the feature level can be a problem. 
For instance, prioritizing features like income level and neighborhood in a loan approval model can inadvertently embed socioeconomic biases, disadvantaging applicants from lower-income areas. This is because these are confounding variables \cite{barriere-cifuentes-2024-text-classifiers,Barriere2024}, creating spurious correlations 
between the target and input variables. These biased features can lead to skewed model predictions, perpetuating inequality

\subsubsection*{Annotation Bias}
Like for data, the annotations can be affected by a clear selection bias. Indeed because of cultural preferences, people will react differently to several subjective phenomena such as hate speech or social acceptability. \citet{Santy2023} show that for these two tasks the annotations vary with respect to the demographics of the annotators. 


\subsubsection*{Cultural bias} 
Data from one culture can be overly represented compared to another one, 
resulting in non-egalitarian behavior of the model. 
First, \citet{Naous2024} showed that LLMs are negatively biases toward Arabic culture: the model will have less cultural references, and be less aligned with human beliefs, norms, and customs from a subordinate cultural group. This is a selection bias. 
Second, an LLM will tend to assimilate cultural stereotypes found in the training data, tending to an amplification of existing cultural prejudices within the model’s outputs. These are group attribution biases.

\subsubsection*{Linguistic biases}
Some languages are prominent in the training set, for example GPT3 has been trained with 50 times more English than French, which is the second language in terms of training data. 
An LLM will confound Chilean and Argentinean Spanish, even though it will not confound Irish and Scottish English. This is due to selection bias, 
as the training data does not represent the real world. But also to representation bias, as the data from the web contains way more Irish/Scottish references than Chilean/Argentinean ones. 

\subsubsection*{Ideological and political biases}
In the training data, some political and ideological biases are more represented than others. \citet{Argyle2023} show that it is possible to replicate the viewpoints of demographically varied U.S. sub-populations by prompting LLMs when prompting the LLM to act like a persona from a specific political side. However, they will tend to favor more certain political perspectives or ideologies, and will be more prone to represent stereotypes of sub-dominant groups \cite{Liu2024}.


\subsubsection*{Demographic biases}
Training data shows an unequal (or lack of) representation of certain demographic groups. This can take the form of \textbf{Geographical Biases}: several works \citep{manvi2024large,dunn2024pre,Godey2024} pointed out that LLM showed a poor geographical knowledge about some parts of the world. It can also be more targeted on the individuals, like a \textbf{Social Class Bias}, as \citet{Curry2024} show that LLMs disadvantage less-privileged socioeconomic groups. 

\begin{figure}
    \centering
    \includegraphics[width=.5\textwidth]{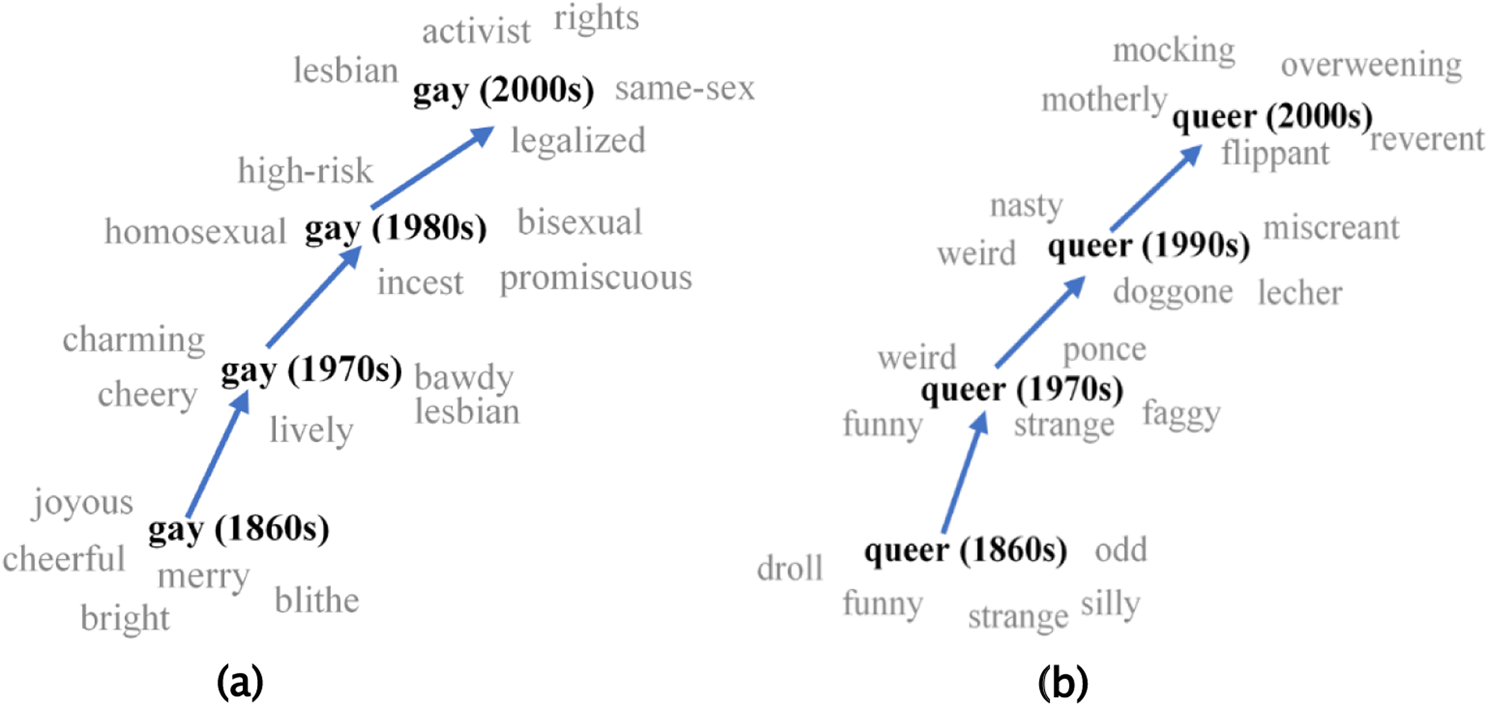}
    \caption{Evolution of meaning of the words \textit{gay} and \textit{queer} accross time \cite{shi2020evolution}}
    \label{fig:lgbt}
\end{figure}

\subsubsection*{Temporal biases}
As the data is selected over a specific time-period, it limits the historical contexts when reporting current events, but also trends or opinions. First the semantic meanings of the words change over time \citep{Schlechtweg2019} (see Figure \ref{fig:lgbt}), but it impacts the results even in a short period \citep{Loureiro2022}. 
Who would like a LLM having the mainstream early 60's vision of women? Or one having the mainstream opinion of slavery that people had in the Eighteenth century? 


\subsubsection*{Confirmation biases} LLM are trained to be aligned with the users beliefs, and will tend to be more assertive about assertive training data such as strong opinions. Also, as they want to satisfy the user, LLM can tend to selective information retention in order to create cognitively appealing instead of informative content

\section{Where to find them?}

Even though they sneak everywhere, there are techniques more or less successful to detect biases in the data, or in the models.

\begin{figure}
    \centering
    \includegraphics[width=.5\textwidth]{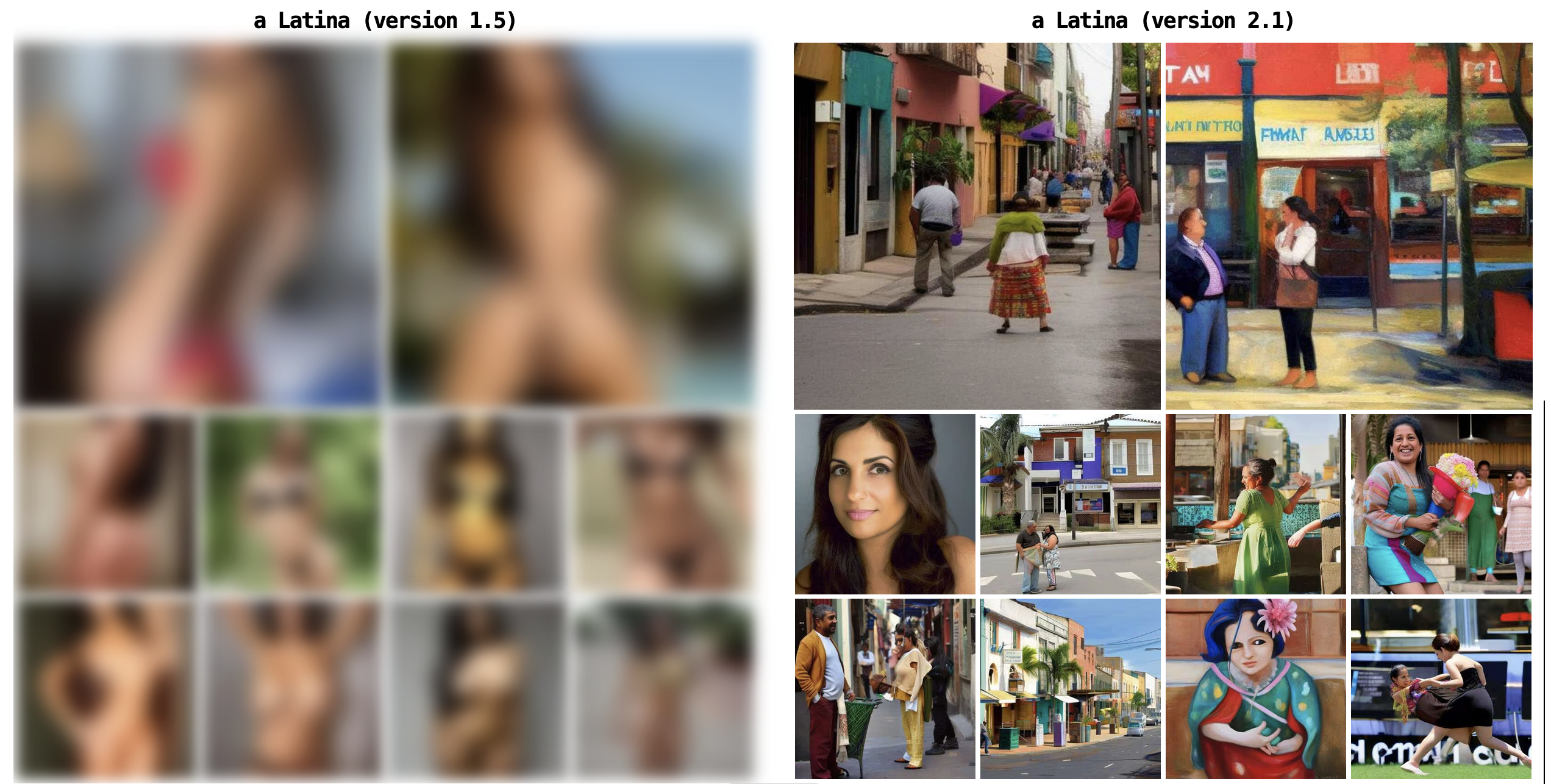}
    \caption{Latina bias of Stable Diffusion version 1.5 compared to version 2.1}
    \label{fig:latina}
\end{figure}

\subsection{In the data}

\subsubsection*{Missing Feature Value} For tabular data, a missing value can be the place of a bias. If some values are missing from a speciific target group, then it may indicate it is under-represented. For example, if the income data for a particular demographic is often missing, the model might misinterpret the economic status of that group.

\subsubsection*{Unexpected Feature Values}
Unexpected feature values could signal data entry errors or other inaccuracies leading to a bias. A negative age for some people might bias the model against certain age groups. 
%
Moreover, identifying unexpected values helps in spotting outliers which may disproportionately represent a minority group. If the model is trained on these outliers without correction, it might generalize poorly on them.

\subsubsection*{Data Skewness} 
If certain groups or characteristics may be under- or over-represented relative to their real-world prevalence, it can introduce a bias into the model. And even if they do represent the real-world prevalence, it might not be the best idea to train on them and repeat real-world inequalities. 
Looking at the data distribution with respect to the different target groups is a good option. It is an easy way to check if there is an important skewness in the represented data between the different target groups. Similarly, the number of samples per target-group is also important to not favor the over-represented group. 

\subsubsection*{Annotation not representative} 
Ask the demographic of the annotators when collecting them is now seeing as essential. \citet{Santy2023} showed the notion of hate speech and social acceptability varies a lot between different cultures, i.e., what might be socially accepted for some individuals might be totally forbidden for others. For this reason, annotations that will be used for supervised learning should not be skewed toward one subgroup of the global population that will be impacted by this model. 

\subsubsection*{Dirty (pre-)training data} 
When looking for collecting a massive dataset in order to pre-train a foundation models, it is likely that the data will not be perfect. Cleaning the dataset in order to remove duplicated text helps to reduce biases, but also to improve the performance of the models: \citet{Hernandez2022a} note that "\textit{for a 1B parameters model, a hundred duplicates are harmful; at 175B, even a few duplicates could have a disproportionate effect}". 
Cleaning by removing hate speech and porn content (which forms the major part of the web) is also a good idea. For example, \citet{Tiku2023} found out that using Stable Diffusion 1.5 version, "latina" produces highly sexual images, as 20\% of the caption containing this word were judge as unsafe by a NSFW classifier (see Figure \ref{fig:latina}). 


\subsection{In the models}

\subsubsection{Positiveness over target groups}
Analyzing the output prediction of a system with respect to different target groups can tell a lot about its functioning, especially when the outputs can be seen as positive or negative. For example, if a sentiment is bad for sentences containing Arabic names, employability lower for female, or recidivism prediction higher for black people, it might not be a good sign. 

\begin{figure}
    \centering
    \includegraphics[width=.5\textwidth]{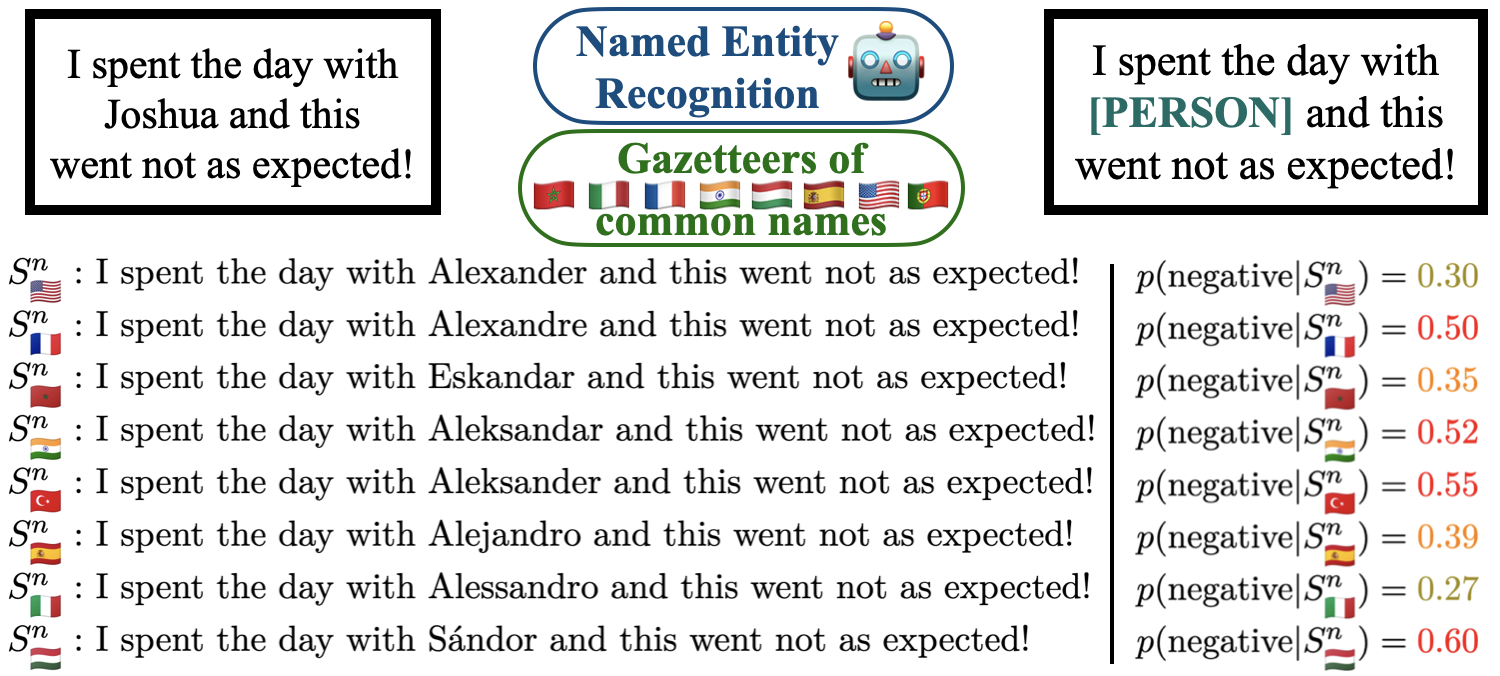}
    \caption{Bias detection with respect to countries, using counterfactual examples 
    over a sentiment analysis system, which should be supposed to output the same predictions.}
    \label{fig:xeno_val}
\end{figure}

\begin{figure*}[ht]
    \centering
    \includegraphics[width=.9\textwidth]{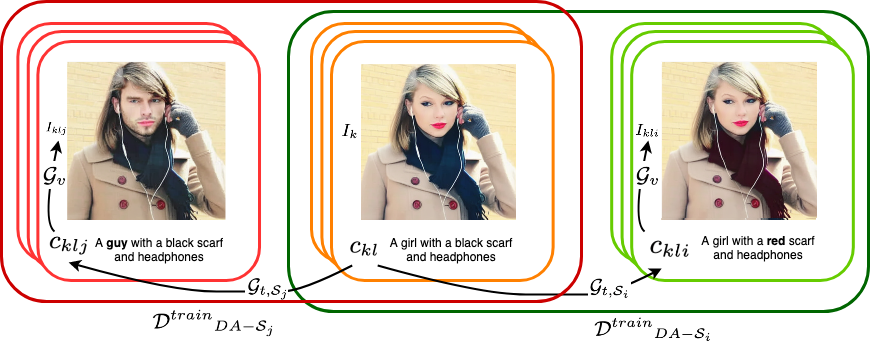}
    \caption{Example of multimodal data-augmentation fostering diversity \cite{barriere-etal-2023-targeted}}
    \label{fig:data_aug}
\end{figure*}

\subsubsection*{Robustness and stability}
In Machine Learning, a bias can be seen as a change of decision influenced by a non-causal variable. \citet{Ribeiroa} use counterfactuals to check the robustness, such as changing a few words or attribute in a sentence and see whether it impacts the model behavior. 
If the model output labels specific to being positive or negative, a bias can be inferred regarding the changes of attributes. Indeed, by using names as proxy it is possible to estimate bias of a model regarding the provenance of a name (\citet{barriere-cifuentes-2024-text-classifiers,Barriere2024}; see Figure \ref{fig:xeno_val}).


\subsubsection*{Heterogeneous performances over target groups}
One simple definition of a bias system is that it performs poorly on data from a target group. If a facial recognition system is not able to work on black people then it is biased.
False positive rate and false negative rate per subgroup can help to understand which groups experience disproportionately worse or better performance. 
This hidden bias can be detrimental, especially in applications like credit scoring or medical diagnoses. 




\subsubsection*{Real-life crash test}
One of the best tests is to put your system in contact with users. It rarely operates perfectly when applied to real, live data. When an issue occurs, evaluate whether it reflects existing societal disadvantages and determine its influence on impacted people. 

\subsection{How to mitigate them?}

There are many ways to mitigate negative biases of models. Here are a few of them. 

\subsubsection*{Over/Subsampling}
If one target group is under- or over-represented, it might hurt the performance of the model. One simple solution is to process to a sampling. 

\subsubsection*{Weighted Samples}
Another solution would be to weight the loss of the function, simply by the inverse of the proportion of each target group sample (if you have 90\% of group A and 10\% of group B, then you can weight the samples of group A by $\frac{1}{0.9}$ and the ones of group B by $\frac{1}{0.1}$. 

\subsubsection*{Objective function reflecting fairness}
One can also create a function to help reducing the impact of biased samples like a focal loss which lower the impact of the easy samples on the weight update, such as a debiased focal loss \cite{Mahabadi2020} or its unsupervised version from \citet{Orgad2022}. 
The latter one relies on a success detector supposed to predict whether or not the main model, without knowing the task, will success to predict: if it can predict the main model success on one sample, then it might contain bias features and should have reduced weight in the loss.

\subsubsection*{Data augmentation}
\citet{Sharma2020b} propose to create, for every sample containing an attribute of the target group, a new sample having the same features (except the protected attribute(s)) but with the opposite protected attribute value, and the same label.
For example, the sentence "\textit{John is an engineer and loves snowboard}" would become "\textit{Jane is an engineer and loves snowboard}". 
This techniques can also be used to remove (non-)social biases and forcing multimodal models to adapt to less common associations that what is in the initial dataset, such as a blue lemon (see Figure \ref{fig:data_aug} from \citet{barriere-etal-2023-targeted}). 

\subsubsection*{Adversarial loss}
The \textit{fairness through blindness} is a technique used to maximize the classifier’s ability to predict the class, while minimizing an adversary network's ability to predict a protected variable \cite{Elazar2018,Wang2021b}. For example if a high-dimension model representation of a Curriculum Vitae can be used to predict whether or not a person should be employed and at the same time cannot be used to predict the gender of a person, then it should be irrespective of the applicant's gender.    

\subsubsection*{Human Perspectives}
By using different perspectives! Discuss with experts from the domain, social scientists, policy makers, and psychologists in order to have a different point of view on the impact of your work. Some recent articles \cite{Curry2023,National2024} 
propose to redefine the roots of the problem using arguments and approaches far from those classically put forward in computer science, for phenomena such as diversity or empathy \cite{Santy2023,Tafreshi2021}.

\subsubsection*{Machine Perspectives}
You can also integrate the fact that sometimes there is no one answer for a question, and many perspectives are needed, or should be represented in models. This is actually a new research field in NLP \cite{abercrombie2022proceedings,abercrombie2024proceedings}. For example, an LLM can be prompted with demographic to represent diverse perspectives \cite{Hayati2023}. This allows the model to be forced to adapt to different subgroups of the population on which it is used. 


\subsubsection*{Integrating all the labels}
In the same vein, it is possible to not train on a aggregation of annotations representing the ground truth but on the annotation distribution. This allows representing better the annotators of subjective tasks such as hate speech or emotion recognition \cite{Kim2018}.

\subsubsection*{User Alignment}
Techniques used by LLM like Reinforcement Learning on Human Preferences can help to reduce some biases, such as hate speech or offensive content generation. However, it also has been shown to reduce the universality of the model \cite{Sorensen2024}. 

\section{Conclusion} 
Biases are everywhere. They can be useful in the reasoning process as they are structuring the world, and even though they might represent the reality they can be harmful toward subgroups of the population. It is important to first detect their impacts, which is possible using some of the methods we presented in this paper, and second minimize them in order to tend to fairer IA models. This is possible through optimizing specific objectives with loss functions, adversarial learning, or using data-augmentation, etc... This step is essential if we are to fulfill the promises of AI for a fairer world.

\bibliography{JRC}
\bibliographystyle{natbib}

\end{document}